\documentclass[runningheads]{llncs}

\usepackage{eccv}

\usepackage{eccvabbrv}
\usepackage[table]{xcolor}
\usepackage{graphicx}
\usepackage{booktabs}
\usepackage{subcaption}
\usepackage{tikz}
\usetikzlibrary{calc}
\usepackage{pgfplots}
\pgfplotsset{compat=1.18}
\usepgfplotslibrary{groupplots}
\pgfplotsset{compat=1.18}
\usepackage{tabularx}
\usepackage{makecell}
\usepackage{float}

\usepackage[accsupp]{axessibility}  

\usepackage{hyperref}

\usepackage{orcidlink}

\begin{document}

\title{Mixture of Cognitive Experts in Large Vision-Language Models} 

\titlerunning{Mixture of Cognitive Experts in Large Vision-Language Models}

\author{Robert Wijaya\orcidlink{0009-0009-5864-9420} \and
Ngai-Man Cheung\orcidlink{0000-0003-0135-3791}}

\authorrunning{R. Wijaya and N.-M. Cheung}

\institute{Singapore University of Technology and Design (SUTD), Singapore \\ \email{robert\_wijaya@mymail.sutd.edu.sg, ngai-man\_cheung@sutd.edu.sg}}

\maketitle

\begin{abstract}
  Large Vision Language Models (LVLMs) require strong reasoning over both visual and textual inputs. Recent work suggests that cognitive elements, especially diverse representations and metacognition, correlate with better performance. Many required perceptual functions are already provided by specialized domain-specific computer vision models, which can serve as the perceptual subsystem for detecting objects, localizing them, inferring states, recovering spatial layout, and reading text. The key challenge is integrating these multi-encoder experts into a trustworthy, interpretable, and coherent representation that improves verifiability and reduces hallucinations. This is difficult because vision-language questions span different cognitive levels, yet most LVLM pipelines apply the same perception-reasoning routing regardless of the query’s demands. We propose an evidence-driven multimodal reasoning framework that uses a Bloom-inspired taxonomy as a hierarchical reasoning protocol. Our two-stage cognitive verbalization first produces a Literal Evidence Summary by decomposing expert outputs into short, atomic evidence statements, then performs Bloom Verbalization to convert these items into a staged reasoning trace. A lightweight Reasoning Trace Module quantitatively analyzes the trace, making evidence usage and reasoning progression explicit. Experiments show improvements in both perception and reasoning and diagnostics indicate that different queries induce different cognitive entry levels and evidence-use trajectories.
  \keywords{Multimodal Reasoning \and Mixture of Experts \and Cognitive Modeling \and Evidence-driven VLM}
\end{abstract}

\section{Introduction}
\label{sec:intro}
From the earliest days of artificial intelligence, researchers and the public alike have been captivated by the idea of building machines that can think and act like humans \cite{marr2010vision,turing2007computing,anderson1990cognitive,weizenbaum1966eliza}. Today, with the rapid advances of large language models (LLMs) \cite{touvron2023llama,brown2020language,team2024gemini,bi2024deepseek}, this long-standing vision feels closer than ever. The line separating human and machine abilities is rapidly fading \cite{bubeck2023sparks}. These systems are now used routinely in everyday life, and people are increasingly depending on them for decision-making \cite{eloundou2023gpts}, medicine \cite{li2023ethics}, and creative tasks \cite{lu2024ai}.

Large vision-language models (LVLMs), in particular, demand strong reasoning across both visual and textual inputs \cite{liu2023visual,alayrac2022flamingo}. Several studies in cognitive science and machine learning \cite{schulze2025visual,hagendorff2023machine,collins2024building,binz2024centaur,kargupta2025cognitive} suggests that fundamental scene perception emerges from multiple cognitive functions, including detecting objects, localizing them, inferring their states, modeling relationships, extracting spatial scene layout, and capturing non-object cues such as written text. Recent work further suggests that models incorporating cognitive elements tend to correlate with greater task success, where diverse internal representations and metacognitive abilities are particularly critical \cite{kargupta2025cognitive,binz2024centaur}. Luckily, many of these perceptual functions are already well supported by specialized computer vision (CV) models, since most contemporary CV systems are domain-specific and designed to excel at particular tasks \cite{oquab2024dinov2learningrobustvisual,zong2023detrscollaborativehybridassignments,kirillov2023segment,lee2023pix2structscreenshotparsingpretraining,liu2023deplotoneshotvisuallanguage,wei2023varyscalingvisionvocabulary,zhang2025biomedclipmultimodalbiomedicalfoundation}. 

These specialized CV models can be viewed as the perceptual part of a vision-language system, implementing different cognitive functions, from detecting object presence, localizing positions, inferring states, extracting spatial layout, and even interpreting non-object cues such as written text. When an LVLM is equipped with such experts (e.g., OCR, open-vocabulary detection, segmentation, layout/geometry cues), the system gains explicit, fine-grained visual evidence that can be carried forward as grounding for reasoning. In this sense, specialized CV models are not mere add-ons for accuracy; they are a principled way to approximate the division of labor in human perception.

The key challenge, then, is to better leverage these multi-encoder experts and integrate their outputs into a trustworthy, interpretable, and coherent representation so that reasoning becomes not only stronger, but also more verifiable and less prone to hallucinations. We draw inspiration from Bloom's taxonomy of cognitive processes \cite{bloom1956taxonomy}. Vision-language questions span qualitatively different levels of cognition, from \emph{remembering} and \emph{understanding} (e.g., identifying objects, reading text, or describing a scene) to \emph{applying} and \emph{analyzing} (e.g., comparing, and reasoning over spatial or relational structure). However, most LVLM inference pipelines treat these queries uniformly, applying the same perception-reasoning routing regardless the cognitive demands of the question \cite{zhou2025perception,li2025perception}. In addition, much recent work emphasizes evaluation and data-centric approaches \cite{schulze2025visual,binz2024centaur,binz2023using,wijaya2024multimodal}, whereas our work focuses on the architectural design.

Bloom's taxonomy is appealing in this setting because it provides an actionable, human-grounded decomposition of cognition that can be operationalized at inference time. It explicitly frames cognition as a hierarchy of thinking processes that progress from recall to higher-level reasoning. 

To operationalize this idea, we propose an evidence-driven framework that uses routed vision experts for evidence extraction and a Bloom-inspired protocol for cognitive synthesis, explicitly separating perception from higher-level reasoning. Our framework follows an autoregressive vision–language architecture similar to LLaVA \cite{liu2024improved}. Given an image and a query, we first build an evidence store using an LLM-guided router that selects a small subset of relevant vision experts. The selected experts produce structured perceptual outputs, which are normalized into a unified evidence store through atomic verbalization. This Literal Evidence Summary prioritizes fidelity and inspectability: each evidence item is a short, non-inferential statement with an explicit type and grounding reference.

Building on this evidence store, a Bloom verbalizer performs staged cognitive synthesis following Bloom’s taxonomy, generating a Bloom-structured draft answer with step-local outputs and explicit citations to evidence identifiers. The Bloom output is produced as text tokens that can be fed directly to the final LLM together with the image and query. To support quantitative analysis without relying on hidden chain-of-thought, we further introduce a lightweight Reasoning Trace Module that parses the Bloom output and evidence store to recover the query-conditioned starting level, executed Bloom steps, and step-wise cited evidence sets, as well as evidence-type usage patterns. Together, these components provide a principled, operational pathway for integrating specialized CV experts into LVLM inference, while making intermediate evidence and reasoning progression explicit and measurable.

We evaluate the method on multiple vision-language benchmarks, with a particular focus on visual perception and reasoning. The resulting model achieves strong results across tests, especially on perception- and reasoning-intensive subsets and improves robustness on hallucination-focused evaluations.

Moreover, the trace analysis provides quantitative evidence that different queries elicit different reasoning trajectories. Specifically, queries demand different cognitive depths, reflected by the entry-level distribution inferred by the Reasoning Trace Module; across our evaluated sets, many queries primarily require recall or understanding (e.g., identifying visible entities or reading text) rather than higher-order analytical reasoning. 

Our contributions are threefold:
\begin{itemize}
    \item We introduce an evidence driven LVLM framework that integrates routed vision experts with a Bloom-inspired, hierarchical reasoning protocol at inference time.
    \item We propose a two-stage cognitive verbalization (Literal Evidence Summary + Bloom Verbalization) that makes evidence grounding explicit and inspectable.
    \item We empirically demonstrate improved performance across a range of vision-language benchmarks, with notable gains on perception- and reasoning intensive evaluations.
\end{itemize}







\section{Related Works}

\subsection{Mixture of Experts}
Jacobs \textit{et al.} \cite{jacobs1991adaptive} first formulated adaptive mixtures of local experts and has been systematized by later surveys \cite{yuksel2012twenty}, while modern sparse MoE implementations \cite{fedus2022switch} demonstrate strong scaling behavior and efficiency at very large parameters counts; more recently, soft MoE \cite{puigcerver2023sparse} aim to mitigate hard-routing instability and improve expert utilization. MoE in VLMs extends the same principle to multimodal learning, where expert specialization can align with modality or semantics: vision-side sparse expert backbones such as V-MoE \cite{riquelme2021scaling} demonstrate efficient scaling for visual representations, and subsequent work scales vision-language training with sparse MoE routing \cite{shen2023scaling} and develops unified pretraining designs that explicitly mix modality-specific and shared pathways, e.g., mixture-of-modality-experts in VLMo \cite{bao2022vlmo}. Several works then instantiate MoE-style modularity with different integration choice: MoAI \cite{lee2024moai} adopts Flamingo-like interface, whereas MoVA \cite{zong2024mova} aligns more closely with the LLaVA-style pipeline, where both utilizing different specialized CV models and report substantial gains especially on the cognition focused subsets, complementing sparse MoE effort such as MoE-LLAVA \cite{lin2026moe}. 

\subsection{LLMs and LVLMs of Human Cognition}
A growing line of work treats foundation models as objects of cognitive science, using controlled paradigms to characterize whether their behaviors resemble human cognition: \cite{binz2023using} evaluates LLMs on classical cognitive tasks (decision-making, search, deliberation, causality), while \cite{hagendorff2023machine} introduces a methodology for probing emerging capabilities beyond standard benchmarks. Moving from measurement to modeling, \cite{binz2023turning} shows that aligning LLMs to behavioral datasets improves their ability to predict human choices; in parallel, \cite{collins2024building} motivates human-compatible thought partners, and behavioral interaction studies such as \cite{akata2025playing} probe social cognition via repeated-game dynamics. For vision-language cognition, supervision designs matters: \cite{doveh2023dense} demonstrate that denser, better-aligned captions can directly strenghten compositional reasoning. \cite{schulze2025visual} tackles the fundamental problem of testing and evaluating whether VLMs can perform cognitive work, and \cite{binz2024centaur} complements this direction with data-centric approach that scales behavioral supervision to model cognition broadly; by contrast, our work aims to improving cognitive ability from the architecture side.

\section{Methodology}
\subsection{Architecture}
\subsubsection{VLM Backbone.} We adopt a decoder-only LLM-based LVLM backbone built on the open-source LLaVA codebase \cite{liu2023visual}, which couples a pretrain vision encoder with an autoregressive language model for multimodal generation. For visual encoding, we use CLIP-L (336 px) \cite{radford2021learning} as the base image tower, and instantiate the language backbone of Vicuna-7B \cite{vicuna2023}.
\subsubsection{Vision Experts Pool.}
\label{sec:visionpool}
Our pipeline requires a VLM with a pool of task-specialized vision experts. In particular, we utilized vision experts from \cite{zong2024mova}. DETR \cite{carion2020end} is used to produce object-centric bounding boxes and labels, and SAM \cite{kirillov2023segment} to refine selected regions into instance masks for accurate spatial predicates and boundary-sensitive queries. For non-object cues, we include Pix2Struct \cite{lee2023pix2structscreenshotparsingpretraining}, Deplot \cite{liu2023deplotoneshotvisuallanguage}, and Vary \cite{wei2023varyscalingvisionvocabulary} as experts for OCR, chart parsing, and document understanding, respectively.

\begin{figure}[tb]
  \centering
  \includegraphics[height=7cm]{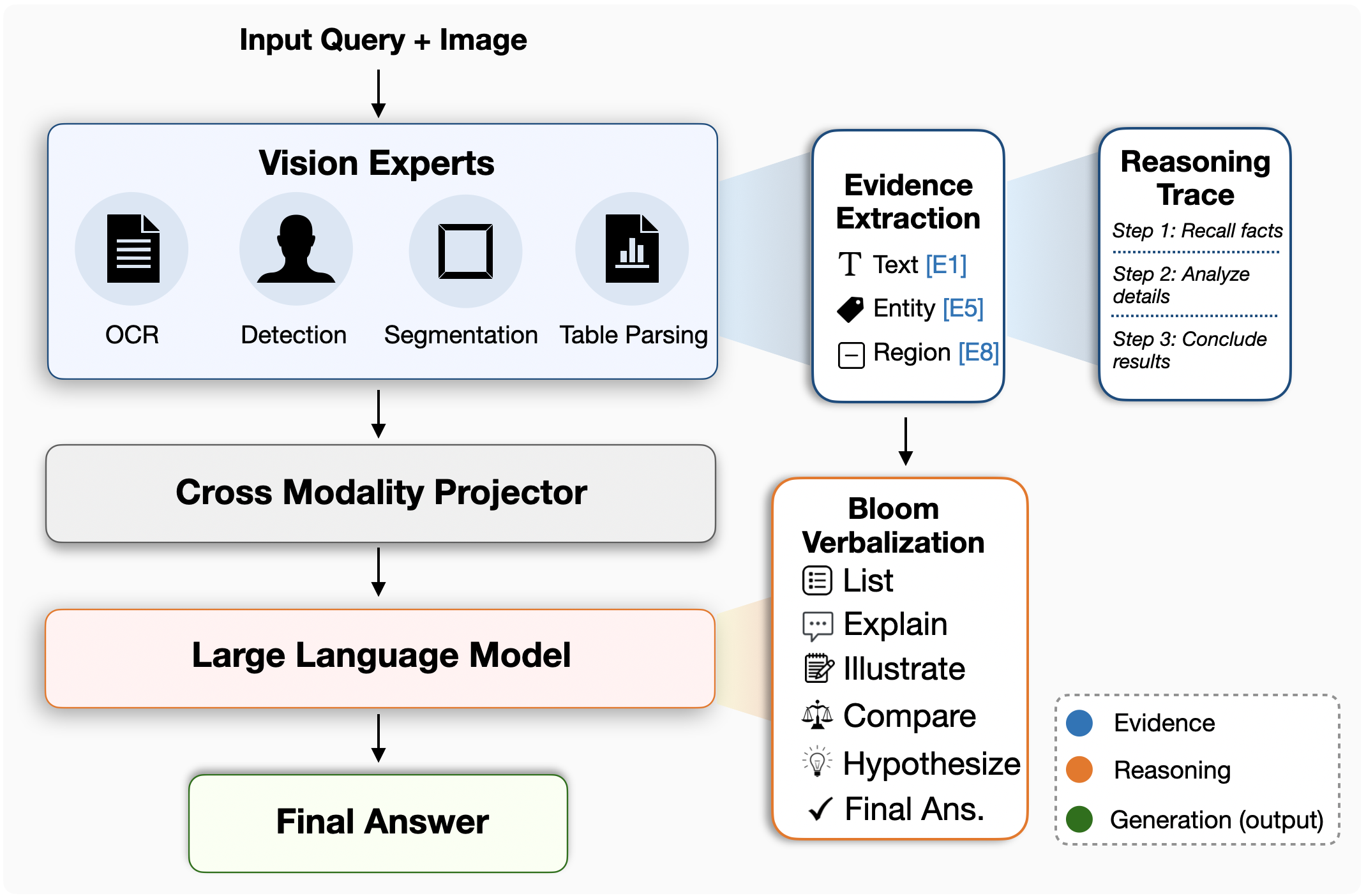}
  \caption{Our method use expert outputs that converted into a Literal Evidence Summary, a collection of atomic, citeable evidence statements indexed in a unified evidence store.  A Bloom-inspired verbalisation processes this evidence store to produce a staged reasoning trace with explicit evidence citations, which is analysed by the Reasoning Trace Module and finally provided to the LLM to generate the final answer.
  }
  \label{fig:example}
\end{figure}
\subsection{Two-stage Cognitive Verbalization}
\subsubsection{Literal Evidence Summary.} Given an input query $q$, the router selects a subset of vision experts indexed by $\mathcal{G}$. The expert pool then produces raw, structured perceptual outputs
\begin{equation}
\mathcal{O}=\{O^{(e)}\}_{e\in\mathcal{G}},
\end{equation}
where each $O^{(e)}$ may contain modality-specific observations such as OCR text lines with bounding boxes, detection labels with boxes, segmentation masks, or parsed tables/charts. This stage takes $(q,\mathcal{O})$ as input and converts these heterogeneous expert outputs into a unified evidence set $\mathcal{E}$, with the explicit goal of capturing \emph{literal} perceptual facts only; no reasoning, speculation, or task-level inference is performed at this stage.

Concretely, we perform \emph{atomic verbalization}: instead of generating a single long description, each expert output is decomposed into many short evidence statements.
For each expert $e$, we define a verbalizer
\begin{equation}
v_e: O^{(e)} \rightarrow \{(s_k,m_k)\}_{k=1}^{n_e},
\end{equation}
where $s_k$ is a concise literal statement and $m_k$ stores grounding metadata (e.g., bounding box coordinates, mask identifiers, or table-cell indices). All statements from the selected experts are then normalized into a unified evidence store
\begin{equation}
\mathcal{E}=\{e_i\}_{i=1}^{M},
\qquad
e_i=(\mathrm{id}_i,\mathrm{type}_i,\mathrm{text}_i,\mathrm{ref}_i).
\end{equation}
where $\mathrm{id}_i$ is a unique identifier used for citation in the subsequent Bloom verbalization stage,
$\mathrm{type}_i \in \{\texttt{TEXT},\texttt{ENTITY},\texttt{REGION}\}$ denotes the evidence category,
$\mathrm{text}_i$ is an atomic \emph{literal} statement, and $\mathrm{ref}_i$ is a grounding pointer linking the statement to its source (e.g., bbox/mask/table-cell reference).

This evidence typing aligns with the functional roles of the underlying experts.
\texttt{TEXT} evidence arises from OCR/document understanding (e.g., Pix2Struct, Vary) when reading is necessary;
\texttt{ENTITY} evidence comes from open-vocabulary detection (e.g., DETR) when object presence is central;
\texttt{REGION} evidence is produced by segmentation/instance-mask models (e.g., SAM) when pixel-accurate localization matters; and
\texttt{TABLE} (and optionally \texttt{CHART}) evidence is extracted by structured parsers (e.g., DePlot, Pix2Struct) when grid or axis structure is key.
Importantly, these categories are not mutually exclusive; a single image can yield multiple evidence types, so $\mathcal{E}$ is designed to preserve complementary perceptual facts that will later be selectively used for systematic reasoning in the next stage.
\begin{figure}[tb]
  \centering
  \includegraphics[height=7cm]{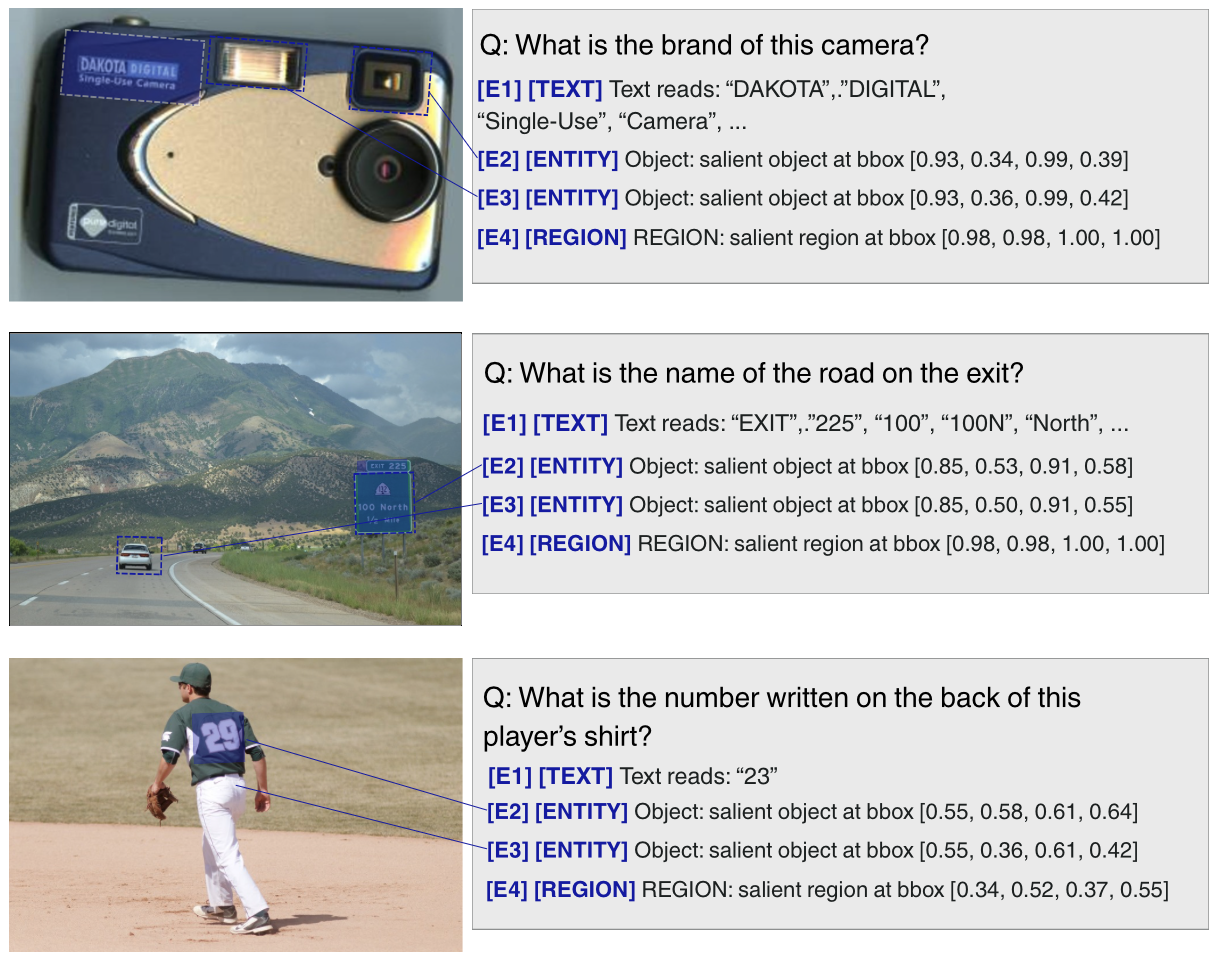}
  \caption{Three qualitative examples of the proposed Literal Evidence Summary produced by specialized vision experts that can be referenced in downstream reasoning.}
  \label{fig:example}
\end{figure}

\subsubsection{Bloom Verbalization.}
\label{bloom-verbalization}
The Bloom verbalizer transforms multiple evidence items into a systematic, staged reasoning trace that follows Bloom’s cognitive hierarchy.
It does not directly act as the final answerer; instead, it produces a Bloom-structured answer as text tokens, which is then fed directly to the LLM.
Given the query $q$ and evidence set $\mathcal{E}$, the Bloom verbalizer produces a comprehensive, evidence-grounded draft answer in a staged manner following Bloom's cognitive hierarchy.
Formally,
\begin{equation}
y_{\mathrm{Bloom}}, \tilde{\mathcal{T}} = f_{\mathrm{Bloom}}(q,\mathcal{E}),
\label{eq:bloom}
\end{equation}
where $y_{\mathrm{Bloom}}$ is the Bloom-structured draft answer and $\tilde{\mathcal{T}}$ denotes the optional step-wise trace retained for the Reasoning Structure Module (Section ).
Unlike the final answerer, the Bloom verbalizer is used to generate an intermediate textual scaffold that can be directly consumed by the downstream LLM as additional context. We define an ordered set of Bloom steps:\\
$\mathcal{L}=[\texttt{LIST},\texttt{EXPLAIN},\texttt{ILLUSTRATE},\texttt{COMPARE},\texttt{HYPOTHESIZE},\texttt{FINAL ANS.}]$.
The query determines a starting level $\ell_0\in\mathcal{L}$; once $\ell_0$ is selected, the verbalizer executes all subsequent steps in the hierarchy until \texttt{FINAL ANS}.
Formally, if $\ell_0=\mathcal{L}[k]$, the executed step set is
$\mathcal{L}(q)=\{\mathcal{L}[k],\mathcal{L}[k+1],\dots,\texttt{FINAL ANS.}\}$.
At each executed step $\ell$, the verbalizer outputs a short step-local text $c_\ell$ grounded by explicit citations to evidence IDs $U_\ell\subseteq\{1,\dots,M\}$.
The concatenation of these step outputs constitutes the Bloom draft answer $y_{\mathrm{Bloom}}$, which is fed to the LLM together with the image $I$ and query $q$:
\begin{equation}
\hat{y} = f_{\mathrm{LLM}}(q, I, y_{\mathrm{Bloom}}).
\label{eq:final}
\end{equation}

A simple example of Bloom verbalization is illustrated in Fig.~\ref{fig:example}. The figure shows how Bloom verbalization converts a set of typed, citeable evidence items into a structured reasoning progression from low- to high-level cognition. Given the question and the evidence store (e.g., OCR text and localized entity/region cues with IDs such as \texttt{[E1]}), the process begins with \texttt{LIST}, which enumerates only literal facts extracted from the evidence (here, the visible tokens ``DAKOTA'' and ``DIGITAL''). It then \texttt{EXPLAIN}s why those facts are relevant to the query (brand identification is typically supported by printed text), and \texttt{ILLUSTRATE}s a grounded example by forming the salient brand phrase ``DAKOTA DIGITAL'' directly from the cited OCR evidence. Next, \texttt{COMPARE} contrasts alternative interpretations (using ``DAKOTA'' alone versus ``DAKOTA DIGITAL'' as the full brand name), before \texttt{HYPOTHESIZE} commits to the best-supported interpretation. Finally, the \texttt{FINAL} step outputs the answer (``Dakota Digital'').

\begin{figure}[tb]
  \centering
  \includegraphics[width=\linewidth, height=9cm, keepaspectratio]{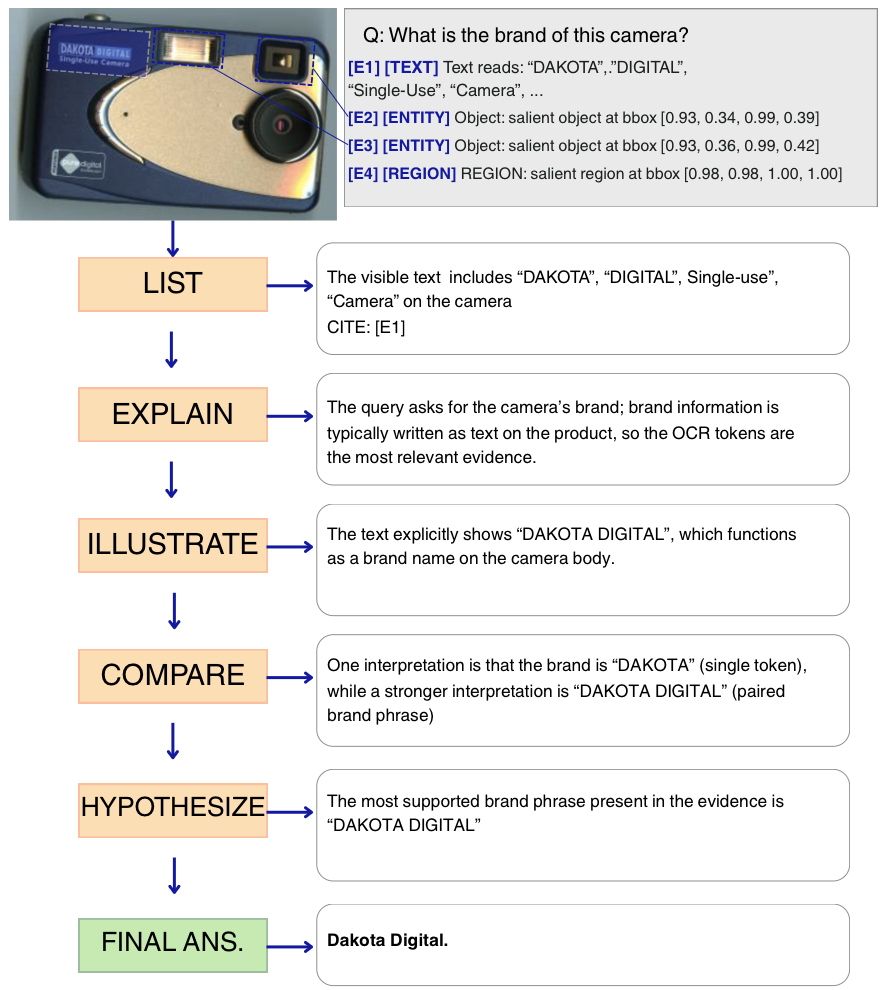}
  \caption{One simple representation to illustrate the Bloom verbalization (LIST to Final Answer) reasoning progression for identifying the brand of a camera.}
  \label{fig:example}
\end{figure}

\subsection{Reasoning Trace Module}
The Reasoning Trace Module is a lightweight, deterministic component that enables fine-grained, quantitative analysis of the Bloom verbalizer's output without requiring access to any hidden chain-of-thought.
Its primary inputs are: (i) $y_{\mathrm{Bloom}}$, a Bloom-structured text that explicitly contains step headers (e.g., \texttt{LIST}, \texttt{EXPLAIN}, $\ldots$, \texttt{FINAL ANS.}) and evidence citations (e.g., \texttt{[E12]}), and (ii) the evidence store $\mathcal{E}=\{e_i\}_{i=1}^{M}$, where each evidence item minimally provides an identifier $\mathrm{id}_i$ (such as \texttt{E12}) and a semantic type $\mathrm{type}_i \in \{\texttt{TEXT}, \texttt{ENTITY}, \texttt{REGION}\}$.
From these inputs, the module produces a single structured trace record that includes the detected starting level $\ell_0$, the executed step set $\mathcal{L}(q)$ implied by the Bloom hierarchy, the per-step cited evidence sets $\{U_\ell\}_{\ell \in \mathcal{L}(q)}$, and the evidence-type mix at each step obtained by mapping cited identifiers through $\mathcal{E}$.
This trace record supports fine-grained quantitative analyses of how evidence is selected, propagated across stages, and ultimately used to justify the final answer.

This module deterministically parses step headers and evidence citations from $(y_{\mathrm{Bloom}},\mathcal{E})$, it enables dataset-level diagnostics of reasoning behavior. In particular, by aggregating the extracted trace records across examples, we can compute summary statistics such as the fraction of outputs that contain each Bloom step, the fraction of those steps that cite at least one evidence ID, and the average number of cited evidence IDs per step $|U_\ell|$ (computed over cited steps only), as well as the distribution of the query-conditioned entry level $\ell_0$. These aggregated metrics directly produce the Bloom-trace diagnostic table shown in Table~\ref{tab:trace-diagnostics}, making reasoning progression and evidence grounding measurable.
\subsection{Router}
The Router is an LLM-guided expert selector that leverages the LLM’s instruction-following (``tool-use'') capability to choose the top-$K$ vision experts most relevant to a given input. Concretely, we form a routing prompt that includes (i) the user query $q$, (ii) brief natural-language descriptions of each candidate expert (e.g., OCR, detection, segmentation, chart parsing), and (iii) a coarse image representation obtained by downsampling the base vision encoder’s visual tokens (routing does not require fine image details, so a compact embedding suffices). The LLM then generates a short textual selection (e.g., expert option letters), which we parse into an index set $\mathcal{G}$ of the selected top-$K$ experts; only experts in $\mathcal{G}$ are executed in the subsequent expert pool stage for fine-grained evidence extraction.

\subsection{Adapter}
Since the output of our Bloom Verbalization (Sec \ref{bloom-verbalization}) is a \emph{textual} draft answer $y_{\mathrm{Bloom}}$ that can be appended as tokens and fed directly to the LLM, we focus the adapter on consolidating visual expert information.
Accordingly, we adopt the adapter from \cite{zong2024mova} to fuse the routed vision-expert representations into a unified, LLM-compatible visual representation.
The adapter consists of $L$ stacked blocks that iteratively refine the base image feature $X$.
We set $X_1=X\in\mathbb{R}^{C\times H\times W}$, and for each block $i$, we take the routed top-$K$ expert features $\{F_j \mid j\in\mathcal{G}\}$, align their spatial resolution to $X_i$ (e.g., via bilinear interpolation).
The fused representation is then processed by a transformer sub-block to yield the next state $X_{i+1}$, enabling repeated expert integration across depth.
After $L$ blocks, we optionally apply a small pooling/residual head to obtain a coarser visual token map and use an MLP projector to match the LLM embedding space, so that the final LLM can attend to a single expert-enriched visual representation together with the query and the appended Bloom draft tokens $y_{\mathrm{Bloom}}$.


\section{Experiments}
\label{sec:blind} 
\subsection{Implementation Details}
Our framework is designed to be \emph{training-free}: all components are executed at inference time, and we do not perform any additional pretraining or finetuning. Given an input image $I$ and query $q$, the pipeline runs (i) an LLM-guided router to select the relevant vision experts, (ii) expert inference to produce structured perceptual outputs, (iii) deterministic evidence construction via template-based verbalization into a unified evidence store $\mathcal{E}$, (iv) Bloom verbalization to generate a staged, evidence-cited draft answer $y_{\mathrm{Bloom}}$, and (v) final answering by conditioning the downstream LLM on $(q, I, y_{\mathrm{Bloom}})$. In addition, our Reasoning Trace Module parses $y_{\mathrm{Bloom}}$ and $\mathcal{E}$ deterministically to recover step-wise evidence for tracing, interpretability and analysis purpose.

\subsubsection{Baselines.} Our goal is not to merely to achieve a new and better performance via additional training, but to validate the feasibility of the proposed \emph{architecture-level} concept, such as that combining specialized perceptual experts with a Bloom-structured reasoning protocol yields more grounded and analyzable multimodal behavior. We therefore compare against widely used LVLM baselines that represent different design points in the current literature: InstructBLIP~\cite{dai2023instructblip} as an instruction-tuned vision--language model, Qwen-VL~\cite{bai2023qwen} as a strong multimodal foundation model with robust OCR and general understanding capabilities, and LLaVA-1.5~\cite{liu2024improved} as a popular open LVLM with competitive performance across general VQA and reasoning benchmarks. All baselines are evaluated under their standard inference settings, and our method is evaluated in a strictly training-free manner to emphasize feasibility and conceptual validation.
InstructBLIP \cite{dai2023instructblip}
Qwen-VL \cite{bai2023qwen} 
LLaVA-1.5 \cite{liu2024improved}

\subsection{VLM Benchmarks}
Our evaluation reports a multi-benchmark comparison of several representative vision–language models (InstructBLIP \cite{dai2023instructblip}, Qwen-VL / Qwen-VL-Chat \cite{bai2023qwen}, LLaVA-1.5 \cite{liu2024improved}) at different parameter scales (7B and 13B). Performance is evaluated on five widely used benchmarks spanning visual question answering and general multimodal reasoning: TextVQA \cite{singh2019towards}, MMBench \cite{liu2024mmbench}, MMVet \cite{yu2023mm}, POPE \cite{li2023evaluating}, and SEED \cite{li2023seed}, allowing a side-by-side view of model capability across complementary test suites.

Overall, our model achieves strong results on all reported benchmarks, with particularly large gains on MMBench and MMVet, while also improving on TextVQA, POPE, and SEED. The table also highlights that increasing parameter count (e.g., LLaVA-1.5 from 7B to 13B) can yield incremental improvements, but the model configuration still outperforms several larger baselines, suggesting that architectural design and expert integration can be more impactful than scale alone in this setting.

\begin{table}[t]
\centering
\setlength{\tabcolsep}{6pt}
\renewcommand{\arraystretch}{1.2}
\begin{tabular}{lllllll}
\toprule
VLMs & Params & TextVQA & MMBench & MMVet & POPE & SEED \\
\midrule
InstructBLIP \cite{dai2023instructblip} & 7B & 50.1 & 36.0 & 26.2 & 76.7 &  58.8\\
InstructBLIP \cite{dai2023instructblip} & 13B & 50.7 & - & 25.6 & 78.9 & - \\
Qwen-VL \cite{bai2023qwen} & 7B & 63.8 & 38.2  & - &  83.6 & 62.3 \\
Qwen-VL-Chat \cite{bai2023qwen} & 7B & 61.5 & 60.6 & - & - & 65.4 \\
LLaVA-1.5 \cite{liu2024improved} & 7B & 58.2 & 64.3 & 30.5 & 85.9 &  66.1\\
LLaVA-1.5 \cite{liu2024improved} & 13B & 61.3 & 67.7 & 35.4 & 85.9 &  68.2\\
LLaVA-NeXT \cite{li2024llavanextinterleavetacklingmultiimagevideo} & 7B & 64.9 & 67.4 & 43.9 & 86.5 &  70.2\\
\rowcolor{gray!15}
MoCE & 7B & \textbf{67.8} & \textbf{75.7} &  \textbf{49.0} & \textbf{87.8} & \textbf{72.6} \\
\bottomrule
\end{tabular}
\caption{The table summarizes evaluation results across TextVQA \cite{singh2019towards}, MMBench \cite{liu2024mmbench}, MMVet \cite{yu2023mm}, POPE \cite{li2023evaluating}, and SEED \cite{li2023seed}for several LVLMs.}
\label{tab:8x7}
\end{table}

\begin{figure*}[t]
\label{interpretability-1}
\centering

\begin{tikzpicture}
\pgfplotbarwidth=1.2pt
\begin{groupplot}[
    group style={group size=4 by 1, horizontal sep=0.3cm},
    xbar,
    width=0.31\textwidth,
    height=4cm,
    xmin=0, xmax=100,
    xtick={0,25,50,75,100},
    grid=major,
    minor tick num=0,
    enlarge y limits=0.18,
    ytick style={draw=none},
    y dir=reverse,
    tick label style={font=\footnotesize},
    label style={font=\footnotesize},
    title style={font=\footnotesize},
    nodes near coords,
    every node near coord/.append style={font=\tiny},
]

\nextgroupplot[
    title={InstructBLIP},
    xlabel={Acc. (\%)},
    symbolic y coords={CP,LR,RR,FP-S},
    ytick=data
]
\addplot coordinates {(46.2,CP) (8.7,LR) (31.0,RR) (39.0,FP-S)};

\nextgroupplot[
    title={Qwen-VL-Chat},
    symbolic y coords={CP,LR,RR,FP-S},
    yticklabels={,,,},
    xlabel={}
]
\addplot coordinates {(68.5,CP) (37.0,LR) (45.7,RR) (67.7,FP-S)};

\nextgroupplot[
    title={LLaVA1.5},
    symbolic y coords={CP,LR,RR,FP-S},
    yticklabels={,,,},
    xlabel={}
]
\addplot coordinates {(70.0,CP) (33.2,LR) (56.2,RR) (68.0,FP-S)};

\nextgroupplot[
    title={MoCE},
    symbolic y coords={CP,LR,RR,FP-S},
    yticklabels={,,,},
    xlabel={}
]
\addplot coordinates {(85.8,CP) (50.0,LR) (78.3,RR) (79.5,FP-S)};

\end{groupplot}
\end{tikzpicture}

\caption{Course Perception (CP), Logical Reasoning (LR), Relation Reasoning (RR), Fine-grained Perception (Single Instance) (FP-S)}
\label{fig:mme_mmb_1x4_stacked}
\end{figure*}
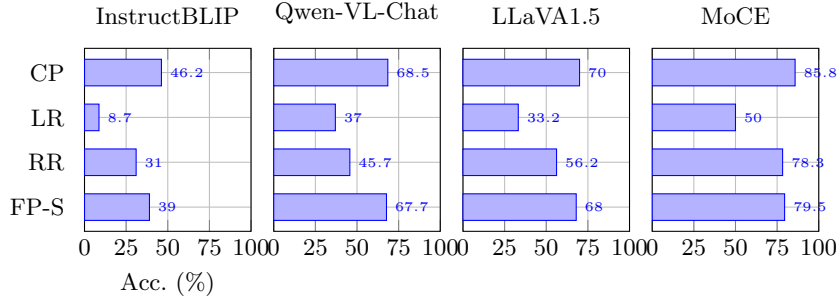

\subsubsection{Perception and Reasoning Evaluation.} As shown in Fig.~\ref{fig:mme_mmb_1x4_stacked}, we break down model accuracy across four MMBench Level-2 abilities: Coarse Perception (CP) and Fine-grained Perception (Single-instance, FP-S) (perception-side), and Logical Reasoning (LR) and Relation Reasoning (RR) (reasoning-side). MMBench is explicitly designed to go beyond a single aggregate score by organizing multimodal evaluation into a hierarchical ability taxonomy (L-1 Perception/Reasoning, then L-2 abilities such as CP, FP-S, and reasoning abilities like LR/RR), enabling diagnosis of which capabilities drive or limit performance.

Across the four compared LVLMs, the bars show distinct “ability profiles”: some models are stronger on coarse scene-level recognition (CP) but weaker on logic or relations, while others improve more uniformly. In particular, our model excel in all four categories in this plot, indicating improvements not only in perception—especially the fine-grained, single-instance regime (FP-S), which is often tied to precise local evidence such as localization/attributes/OCR, but also in higher-level reasoning dimensions (LR and RR) that test structured and relational understanding.

In addition, the plots in Fig. \ref{fig:mme_mmb_1x4_stacked} show that our model consistently attains  perceptual attributes and OCR-related skills, as well as broader scene understanding facets such as scene/category recognition and attribute-centric evaluation. In particular, our model with the proposed method improves both fine-grained perception signals (e.g., OCR/attributes) and more reasoning-oriented dimensions (e.g., relations/spatial), indicating stronger cross-modal integration rather than a narrow specialization.

The bottom row further suggests robustness and reliability improvements. On POPE, which is designed to probe object hallucination and faithfulness through different sampling settings, our model able to maintains the strong performance with comparatively smaller degradation under adversarial conditions. This implies that the predictions are not only more accurate but also more stable when the evaluation is constructed to expose spurious correlations or overconfident errors. On SEED, the model consistently retain good accuracy across multiple scene-level categories (scene, identity, attributes, location, count) indicates more comprehensive scene understanding that generalizes beyond a single dataset style.

These results imply that the architecture provides a more balanced capability profile: it strengthens fine-grained perceptual grounding, supports spatial and relational reasoning, and improves robustness against hallucination-focused stress tests. In practical terms, it appears better at extracting the right visual evidence and using it coherently across tasks, which is aligned with the intended benefit of expert-centric or compositional multimodal designs, specifically, to improve grounding and reasoning fidelity in diverse real-world settings.






\begin{figure*}[t]
\label{interpretability-2}
\centering
\begin{tikzpicture}

\begin{groupplot}[
    group style={group size=2 by 2, horizontal sep=1.2cm, vertical sep=1.2cm},
    width=0.48\textwidth,
    height=4cm,
    grid=major,
    tick label style={font=\footnotesize},
    label style={font=\footnotesize},
]

\nextgroupplot[
    xlabel={},
    ylabel={Acc. (\%)},
    ymin=0, ymax=100,
    symbolic x coords={Attribute,Localization,OCR,Relation},
    xtick=data,
    x tick label style={rotate=25, anchor=east},
]
\addplot+[mark=*] coordinates {(Attribute,51.6) (Localization,3.7) (OCR,21.7) (Relation,17.1)};
\addplot+[mark=*] coordinates {(Attribute,71.9) (Localization,9.7) (OCR,83.4) (Relation,8.3)};
\addplot+[mark=*] coordinates {(Attribute,80.2) (Localization,42.4) (OCR,80.7) (Relation,25.8)};
\addplot+[mark=*] coordinates {(Attribute,89.1) (Localization,62.9) (OCR,84.6) (Relation,53.3)};

\nextgroupplot[
    xlabel={},
    ylabel={Score},
    ymin=0, ymax=60,
    symbolic x coords={Recognition,Spatial,OCR},
    xtick=data,
    x tick label style={rotate=25, anchor=east},
    legend to name=sharedlegend,
    legend columns=4,
    legend style={draw=none, font=\footnotesize},
]
\addplot+[mark=*] coordinates {(Recognition,26.2) (Spatial,13.0) (OCR,19.6)};
\addlegendentry{InstructBLIP}
\addplot+[mark=*] coordinates {(Recognition,33.5) (Spatial,26.2) (OCR,31.8)};
\addlegendentry{Qwen-VL}
\addplot+[mark=*] coordinates {(Recognition,36.1) (Spatial,23.1) (OCR,27.4)};
\addlegendentry{LLaVA1.5}
\addplot+[mark=*] coordinates {(Recognition,46.7) (Spatial,54.0) (OCR,52.2)};
\addlegendentry{MoCE}


\nextgroupplot[
    xlabel={},
    ylabel={Score},
    ymin=50, ymax=100, 
    symbolic x coords={Random,Popular,Adversarial},
    xtick=data,
    x tick label style={rotate=25, anchor=east},
    legend to name=sharedlegend,
    legend columns=4,
    legend style={draw=none, font=\footnotesize},
]
\addplot+[mark=*] coordinates {(Random,76.7) (Popular,76.7) (Adversarial,67.4)};
\addlegendentry{InstructBLIP}
\addplot+[mark=*] coordinates {(Random,86.1) (Popular,83.6) (Adversarial,81.1)};
\addlegendentry{Qwen-VL}
\addplot+[mark=*] coordinates {(Random,89.5) (Popular,85.7) (Adversarial,81.9)};
\addlegendentry{LLaVA1.5}
\addplot+[mark=*] coordinates {(Random,88.9) (Popular,87.8) (Adversarial,85.9)};
\addlegendentry{MoCE}

\nextgroupplot[
    xlabel={},
    ylabel={Acc. (\%)},
    ymin=0, ymax=100,
    symbolic x coords={Scene,Identity,Attributes,Location,Count},
    xtick=data,
    x tick label style={rotate=25, anchor=east},
    legend style={draw=none},
    legend entries={},
]
\addplot+[mark=*] coordinates {(Scene,18.5) (Identity,13.2) (Attributes,12.1) (Location,13.2) (Count,13.2)};
\addplot+[mark=*] coordinates {(Scene,66.8) (Identity,59.6) (Attributes,60.0) (Location,44.5) (Count,40.0)};
\addplot+[mark=*] coordinates {(Scene,75.8) (Identity,74.0) (Attributes,74.0) (Location,68.3) (Count,66.8)};
\addplot+[mark=*] coordinates {(Scene,80.2) (Identity,80.5) (Attributes,79.0) (Location,77.3) (Count,70.4)};

\end{groupplot}

\path (current bounding box.south west) -- (current bounding box.south east)
  node[midway, anchor=north, yshift=-6pt] {\pgfplotslegendfromname{sharedlegend}};

\end{tikzpicture}
\caption{Comparing the scores and accuracies of dimensions related to real-world scene understanding.
Upper Row: MMBench \& MM-Vet; Bottom Row: POPE \& SEED.}
\label{fig:2x2_linecharts_dup}
\end{figure*}
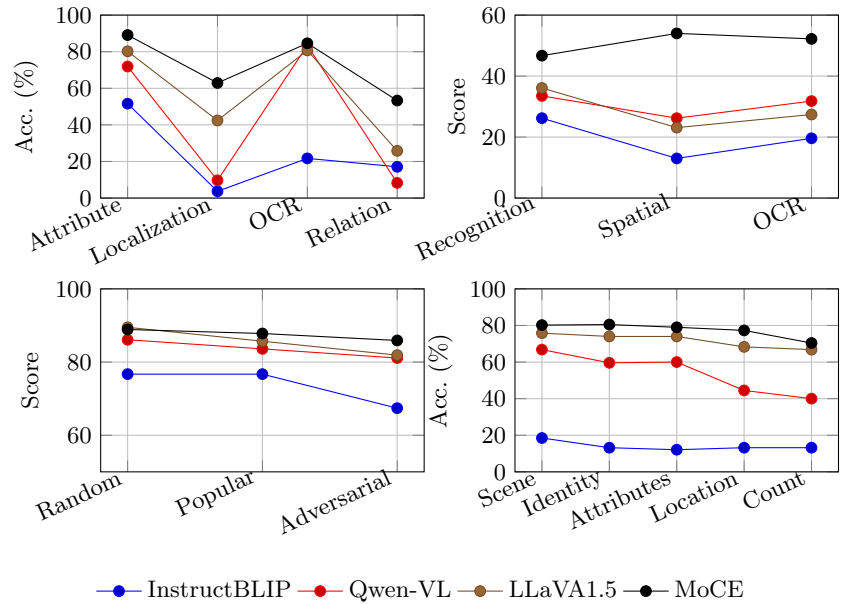

\begin{table}[t]
\centering
\label{tab:vision_experts_bloom}
\small
\setlength{\tabcolsep}{4pt}
\renewcommand{\arraystretch}{0.95}
\begin{tabular}{lccc}
\toprule
\textbf{Metric} &
\makecell{\textbf{Base VLM}} &
\makecell{\textbf{Vision Expert}} &
\makecell{\textbf{VE + Bloom}} \\
\midrule
MMB                   & 64.3 & 70.4        & \textbf{75.7}        \\
MM-Vet                & 31.1 & \textbf{49.3}        & 49.0        \\
SEED Text Understanding & 37.65 & 47.06 & \textbf{51.76}\\
Avg. Latency (s)         & \textbf{3.07} & 10.24  & 13.40    \\
GFLOPs                & \textbf{10,304.91} & 10,982.88  & 12,492.82  \\
Avg. Token Length     & 126.0 & 63.70       & 76.87       \\
\bottomrule
\end{tabular}
\caption{Ablation: Effect of adding vision experts and Bloom verbalization over the base VLM.}
\end{table}

\begin{table}[t]
\centering
\small
\begin{tabular}{lccc}
\toprule
\textbf{Bloom step} & \textbf{Step present (\%)} & \textbf{Step cites evidence (\%)} & \textbf{Avg. $|U_\ell|$ (cited only)} \\
\midrule
\texttt{LIST}        & 15.96 & 71.80 & 1.62 \\
\texttt{EXPLAIN}     & 15.68 & 83.80 & 1.56 \\
\texttt{ILLUSTRATE}  & 13.84 & 87.72 & 1.60 \\
\texttt{COMPARE}     & 15.74 & 83.48 & 1.64 \\
\texttt{HYPOTHESIZE} & 14.10 & 85.11 & 1.53 \\
\texttt{FINAL}       & 14.18 & 79.97 & 1.73 \\
\midrule
\multicolumn{4}{l}{\textbf{Start level distribution} (among detected traces, $N{=}952$)}\\
\texttt{LIST}        & \multicolumn{3}{c}{798 \ \ (83.82\%)}\\
\texttt{COMPARE}     & \multicolumn{3}{c}{79 \ \ (8.30\%)}\\
\texttt{EXPLAIN}     & \multicolumn{3}{c}{46 \ \ (4.83\%)}\\
\texttt{ILLUSTRATE}  & \multicolumn{3}{c}{29 \ \ (3.05\%)}\\
\bottomrule
\end{tabular}
\caption{\textbf{Bloom-trace diagnostics (N=5000).}
We measure (i) the fraction of examples whose Bloom output contains each step header, (ii) the fraction of those steps that include at least one cited evidence ID, and (iii) the average number of cited evidence IDs per step, $|U_\ell|$, computed only over steps with non-empty citations through the Reasoning Trace Module's output. The start-level block summarizes the query-conditioned entry level $\ell_0$ among examples where a start level is detectable.}
\label{tab:trace-diagnostics}
\end{table}

\subsection{Ablation}
\subsubsection{Modality-level Analysis.}
A key concern in evaluating Bloom is the lack of a controlled ablation that isolates the effect of Bloom-guided synthesis from the underlying vision expert pool. To address this, Table~\ref{tab:vision_experts_bloom} reports three settings: the \emph{Base VLM} (no experts, no Bloom), \emph{Vision Experts} (VE), and \emph{VE + Bloom}. We report accuracy on MMBench (MMB), MM-Vet, and SEED Text Understanding, and measure efficiency statistics (latency, GFLOPs, and average token length) on 200 randomly sampled COCO \texttt{val2017} images using the fixed prompt ``Describe this image.''

Relative to the Base VLM, adding vision experts yields the largest jump on perception/grounding-heavy evaluations: MMB improves from 64.3 to 70.4 (+6.1), MM-Vet from 31.1 to 49.3 (+18.2), and SEED Text Understanding from 37.65 to 47.06 (+9.41). This accuracy gain comes with a clear efficiency trade-off: latency increases from 3.07s to 10.24s and GFLOPs from 10,304.91 to 10,982.88, indicating that routing through specialized experts is substantially more expensive than a single-pass base model. Notably, VE also \emph{reduces} average output length (126.0 to 63.70 tokens), suggesting that stronger low-level evidence extraction can make responses more concise even as computation increases.

On top of VE, Bloom consistently improves text-centric and general multimodal performance: MMB increases from 70.4 to 75.7 (+5.3) and SEED text understanding improves from 47.06 to 51.76 (+4.70), while MM-Vet remains comparable (49.3 vs.\ 49.0). The additional synthesis stage incurs a moderate computational cost, increasing end-to-end latency from 10.24s to 13.40s and runtime GFLOPs from 10,982.88 to 12,492.82 (+13.75\%), along with a longer average generation length (63.70 to 76.87 tokens). Overall, the table highlights a three-way trade-off: (i) the Base VLM is fastest/cheapest but least accurate, (ii) VE delivers the biggest accuracy gains (especially on MM-Vet) at a large latency increase, and (iii) Bloom recovers additional gains in structured, text-oriented benchmarks at further compute and modestly longer outputs.

\subsubsection{Step Progresion and Hierarchy Compliance.} We verify that Bloom traces follow the intended step ordering and remain evidence-grounded. Table~\ref{tab:trace-diagnostics} reports, for each step, the frequency of the step header, the fraction of steps that include at least one cited evidence ID, and the average number of cited IDs $|U_\ell|$ (computed over cited steps). Overall, when a step is produced it is usually grounded (\mbox{72\%--88\%} include citations) with a small, consistent support set ($|U_\ell|\approx 1.5$--1.7). The detected entry-level distribution indicates that most traces start at \texttt{LIST} (83.82\%), with smaller fractions starting at higher levels such as \texttt{COMPARE}, \texttt{EXPLAIN}, and \texttt{ILLUSTRATE}.

We further observe that different queries demand different cognitive depths, which is reflected by the start-level distribution inferred by the Reasoning Trace Module. As shown in Table~\ref{tab:trace-diagnostics}, while a non-trivial portion of examples trigger higher entry levels such as \texttt{COMPARE}, \texttt{EXPLAIN}, or \texttt{ILLUSTRATE}, the majority of detectable traces begin at \texttt{LIST} (83.82\%). This suggests that most queries in the evaluated set primarily require basic recall or understanding, such as identifying visible entities or reading text, rather than higher-order analytical reasoning. At the same time, the presence of higher start levels confirms that the framework can adapt its reasoning trajectory to more demanding questions when needed, enabling a principled way to characterize query difficulty and reasoning requirements.

\section{Conclusion}
We propose a framework that integrates specialized vision experts into LVLM reasoning by separating literal evidence extraction from staged cognitive synthesis and making intermediate reasoning traceable. Across multiple evaluations, the resulting model achieves strong performance, with particularly notable improvements on perception and reasoning-intensive evaluations of real-world scene understanding. Our central claim is that imposing an evidence-grounded, hierarchical reasoning protocol yields more verifiable and analyzable multimodal behavior; this is supported by the structured evidence store and the Reasoning Trace Module, which enable step-wise measurement of evidence usage, hierarchy compliance, and query-conditioned cognitive entry levels. These results provide feasibility-level evidence that cognitive-inspired structure can be utilized to improve performance of vision-language models.
\par\vfill\par

\clearpage  


%
%
\bibliographystyle{splncs04}
\bibliography{main}

@book{marr2010vision,
  title={Vision: A computational investigation into the human representation and processing of visual information},
  author={Marr, David},
  year={2010},
  publisher={MIT press}
}

@incollection{turing2007computing,
  title={Computing machinery and intelligence},
  author={Turing, Alan M},
  booktitle={Parsing the Turing test: Philosophical and methodological issues in the quest for the thinking computer},
  pages={23--65},
  year={2007},
  publisher={Springer}
}

@article{anderson1990cognitive,
  title={Cognitive modeling and intelligent tutoring},
  author={Anderson, John R and Boyle, C Franklin and Corbett, Albert T and Lewis, Matthew W},
  journal={Artificial intelligence},
  volume={42},
  number={1},
  pages={7--49},
  year={1990}
}

@article{weizenbaum1966eliza,
  title={ELIZA—a computer program for the study of natural language communication between man and machine},
  author={Weizenbaum, Joseph},
  journal={Communications of the ACM},
  volume={9},
  number={1},
  pages={36--45},
  year={1966},
  publisher={ACM New York, NY, USA}
}

@article{touvron2023llama,
  title={Llama: Open and efficient foundation language models},
  author={Touvron, Hugo and Lavril, Thibaut and Izacard, Gautier and Martinet, Xavier and Lachaux, Marie-Anne and Lacroix, Timoth{\'e}e and Rozi{\`e}re, Baptiste and Goyal, Naman and Hambro, Eric and Azhar, Faisal and others},
  journal={arXiv preprint arXiv:2302.13971},
  year={2023}
}

@article{brown2020language,
  title={Language models are few-shot learners},
  author={Brown, Tom and Mann, Benjamin and Ryder, Nick and Subbiah, Melanie and Kaplan, Jared D and Dhariwal, Prafulla and Neelakantan, Arvind and Shyam, Pranav and Sastry, Girish and Askell, Amanda and others},
  journal={Advances in neural information processing systems},
  volume={33},
  pages={1877--1901},
  year={2020}
}

@article{team2024gemini,
  title={Gemini: A family of highly capable multimodal models, 2024},
  author={Team, Gemini and Anil, R and Borgeaud, S and Wu, Y and Alayrac, JB and Yu, J and Soricut, R and Schalkwyk, J and Dai, AM and Hauth, A and others},
  journal={arXiv preprint arXiv:2312.11805},
  volume={10},
  year={2024}
}

@article{bi2024deepseek,
  title={Deepseek llm: Scaling open-source language models with longtermism},
  author={Bi, Xiao and Chen, Deli and Chen, Guanting and Chen, Shanhuang and Dai, Damai and Deng, Chengqi and Ding, Honghui and Dong, Kai and Du, Qiushi and Fu, Zhe and others},
  journal={arXiv preprint arXiv:2401.02954},
  year={2024}
}

@article{liu2023visual,
  title={Visual instruction tuning},
  author={Liu, Haotian and Li, Chunyuan and Wu, Qingyang and Lee, Yong Jae},
  journal={Advances in neural information processing systems},
  volume={36},
  pages={34892--34916},
  year={2023}
}

@article{alayrac2022flamingo,
  title={Flamingo: a visual language model for few-shot learning},
  author={Alayrac, Jean-Baptiste and Donahue, Jeff and Luc, Pauline and Miech, Antoine and Barr, Iain and Hasson, Yana and Lenc, Karel and Mensch, Arthur and Millican, Katherine and Reynolds, Malcolm and others},
  journal={Advances in neural information processing systems},
  volume={35},
  pages={23716--23736},
  year={2022}
}

@article{jacobs1991adaptive,
  title={Adaptive mixtures of local experts},
  author={Jacobs, Robert A and Jordan, Michael I and Nowlan, Steven J and Hinton, Geoffrey E},
  journal={Neural computation},
  volume={3},
  number={1},
  pages={79--87},
  year={1991},
  publisher={MIT Press}
}

@article{yuksel2012twenty,
  title={Twenty years of mixture of experts},
  author={Yuksel, Seniha Esen and Wilson, Joseph N and Gader, Paul D},
  journal={IEEE transactions on neural networks and learning systems},
  volume={23},
  number={8},
  pages={1177--1193},
  year={2012},
  publisher={IEEE}
}

@article{riquelme2021scaling,
  title={Scaling vision with sparse mixture of experts},
  author={Riquelme, Carlos and Puigcerver, Joan and Mustafa, Basil and Neumann, Maxim and Jenatton, Rodolphe and Susano Pinto, Andr{\'e} and Keysers, Daniel and Houlsby, Neil},
  journal={Advances in Neural Information Processing Systems},
  volume={34},
  pages={8583--8595},
  year={2021}
}

@article{fedus2022switch,
  title={Switch transformers: Scaling to trillion parameter models with simple and efficient sparsity},
  author={Fedus, William and Zoph, Barret and Shazeer, Noam},
  journal={Journal of Machine Learning Research},
  volume={23},
  number={120},
  pages={1--39},
  year={2022}
}

@article{puigcerver2023sparse,
  title={From sparse to soft mixtures of experts},
  author={Puigcerver, Joan and Riquelme, Carlos and Mustafa, Basil and Houlsby, Neil},
  journal={arXiv preprint arXiv:2308.00951},
  year={2023}
}

@article{bao2022vlmo,
  title={Vlmo: Unified vision-language pre-training with mixture-of-modality-experts},
  author={Bao, Hangbo and Wang, Wenhui and Dong, Li and Liu, Qiang and Mohammed, Owais Khan and Aggarwal, Kriti and Som, Subhojit and Piao, Songhao and Wei, Furu},
  journal={Advances in neural information processing systems},
  volume={35},
  pages={32897--32912},
  year={2022}
}

@article{shen2023scaling,
  title={Scaling vision-language models with sparse mixture of experts},
  author={Shen, Sheng and Yao, Zhewei and Li, Chunyuan and Darrell, Trevor and Keutzer, Kurt and He, Yuxiong},
  journal={arXiv preprint arXiv:2303.07226},
  year={2023}
}

@article{lin2026moe,
  title={Moe-llava: Mixture of experts for large vision-language models},
  author={Lin, Bin and Tang, Zhenyu and Ye, Yang and Huang, Jinfa and Zhang, Junwu and Pang, Yatian and Jin, Peng and Ning, Munan and Luo, Jiebo and Yuan, Li},
  journal={IEEE Transactions on Multimedia},
  year={2026},
  publisher={IEEE}
}

@inproceedings{lee2024moai,
  title={Moai: Mixture of all intelligence for large language and vision models},
  author={Lee, Byung-Kwan and Park, Beomchan and Won Kim, Chae and Man Ro, Yong},
  booktitle={European Conference on Computer Vision},
  pages={273--302},
  year={2024},
  organization={Springer}
}

@article{zong2024mova,
  title={Mova: Adapting mixture of vision experts to multimodal context},
  author={Zong, Zhuofan and Ma, Bingqi and Shen, Dazhong and Song, Guanglu and Shao, Hao and Jiang, Dongzhi and Li, Hongsheng and Liu, Yu},
  journal={Advances in Neural Information Processing Systems},
  volume={37},
  pages={103305--103333},
  year={2024}
}

@article{schulze2025visual,
  title={Visual cognition in multimodal large language models},
  author={Schulze Buschoff, Luca M and Akata, Elif and Bethge, Matthias and Schulz, Eric},
  journal={Nature Machine Intelligence},
  volume={7},
  number={1},
  pages={96--106},
  year={2025},
  publisher={Nature Publishing Group UK London}
}

@article{akata2025playing,
  title={Playing repeated games with large language models},
  author={Akata, Elif and Schulz, Lion and Coda-Forno, Julian and Oh, Seong Joon and Bethge, Matthias and Schulz, Eric},
  journal={Nature Human Behaviour},
  pages={1--11},
  year={2025},
  publisher={Nature Publishing Group UK London}
}

@article{binz2023turning,
  title={Turning large language models into cognitive models},
  author={Binz, Marcel and Schulz, Eric},
  journal={arXiv preprint arXiv:2306.03917},
  year={2023}
}

@article{hagendorff2023machine,
  title={Machine psychology},
  author={Hagendorff, Thilo and Dasgupta, Ishita and Binz, Marcel and Chan, Stephanie CY and Lampinen, Andrew and Wang, Jane X and Akata, Zeynep and Schulz, Eric},
  journal={arXiv preprint arXiv:2303.13988},
  year={2023}
}

@article{collins2024building,
  title={Building machines that learn and think with people},
  author={Collins, Katherine M and Sucholutsky, Ilia and Bhatt, Umang and Chandra, Kartik and Wong, Lionel and Lee, Mina and Zhang, Cedegao E and Zhi-Xuan, Tan and Ho, Mark and Mansinghka, Vikash and others},
  journal={Nature human behaviour},
  volume={8},
  number={10},
  pages={1851--1863},
  year={2024},
  publisher={Nature Publishing Group UK London}
}

@article{binz2024centaur,
  title={Centaur: a foundation model of human cognition},
  author={Binz, Marcel and Akata, Elif and Bethge, Matthias and Br{\"a}ndle, Franziska and Callaway, Fred and Coda-Forno, Julian and Dayan, Peter and Demircan, Can and Eckstein, Maria K and {\'E}ltet{\H{o}}, No{\'e}mi and others},
  journal={arXiv preprint arXiv:2410.20268},
  year={2024}
}

@article{doveh2023dense,
  title={Dense and aligned captions (dac) promote compositional reasoning in vl models},
  author={Doveh, Sivan and Arbelle, Assaf and Harary, Sivan and Herzig, Roei and Kim, Donghyun and Cascante-Bonilla, Paola and Alfassy, Amit and Panda, Rameswar and Giryes, Raja and Feris, Rogerio and others},
  journal={Advances in Neural Information Processing Systems},
  volume={36},
  pages={76137--76150},
  year={2023}
}

@article{binz2023using,
  title={Using cognitive psychology to understand GPT-3},
  author={Binz, Marcel and Schulz, Eric},
  journal={Proceedings of the National Academy of Sciences},
  volume={120},
  number={6},
  pages={e2218523120},
  year={2023},
  publisher={National Academy of Sciences}
}

@misc{oquab2024dinov2learningrobustvisual,
      title={DINOv2: Learning Robust Visual Features without Supervision}, 
      author={Maxime Oquab and Timothée Darcet and Théo Moutakanni and Huy Vo and Marc Szafraniec and Vasil Khalidov and Pierre Fernandez and Daniel Haziza and Francisco Massa and Alaaeldin El-Nouby and Mahmoud Assran and Nicolas Ballas and Wojciech Galuba and Russell Howes and Po-Yao Huang and Shang-Wen Li and Ishan Misra and Michael Rabbat and Vasu Sharma and Gabriel Synnaeve and Hu Xu and Hervé Jegou and Julien Mairal and Patrick Labatut and Armand Joulin and Piotr Bojanowski},
      year={2024},
      eprint={2304.07193},
      archivePrefix={arXiv},
      primaryClass={cs.CV},
      url={https://arxiv.org/abs/2304.07193}, 
}

@misc{zong2023detrscollaborativehybridassignments,
      title={DETRs with Collaborative Hybrid Assignments Training}, 
      author={Zhuofan Zong and Guanglu Song and Yu Liu},
      year={2023},
      eprint={2211.12860},
      archivePrefix={arXiv},
      primaryClass={cs.CV},
      url={https://arxiv.org/abs/2211.12860}, 
}

@misc{kirillov2023segment,
      title={Segment Anything}, 
      author={Alexander Kirillov and Eric Mintun and Nikhila Ravi and Hanzi Mao and Chloe Rolland and Laura Gustafson and Tete Xiao and Spencer Whitehead and Alexander C. Berg and Wan-Yen Lo and Piotr Dollár and Ross Girshick},
      year={2023},
      eprint={2304.02643},
      archivePrefix={arXiv},
      primaryClass={cs.CV},
      url={https://arxiv.org/abs/2304.02643}, 
}

@misc{lee2023pix2structscreenshotparsingpretraining,
      title={Pix2Struct: Screenshot Parsing as Pretraining for Visual Language Understanding}, 
      author={Kenton Lee and Mandar Joshi and Iulia Turc and Hexiang Hu and Fangyu Liu and Julian Eisenschlos and Urvashi Khandelwal and Peter Shaw and Ming-Wei Chang and Kristina Toutanova},
      year={2023},
      eprint={2210.03347},
      archivePrefix={arXiv},
      primaryClass={cs.CL},
      url={https://arxiv.org/abs/2210.03347}, 
}

@misc{liu2023deplotoneshotvisuallanguage,
      title={DePlot: One-shot visual language reasoning by plot-to-table translation}, 
      author={Fangyu Liu and Julian Martin Eisenschlos and Francesco Piccinno and Syrine Krichene and Chenxi Pang and Kenton Lee and Mandar Joshi and Wenhu Chen and Nigel Collier and Yasemin Altun},
      year={2023},
      eprint={2212.10505},
      archivePrefix={arXiv},
      primaryClass={cs.CL},
      url={https://arxiv.org/abs/2212.10505}, 
}

@misc{wei2023varyscalingvisionvocabulary,
      title={Vary: Scaling up the Vision Vocabulary for Large Vision-Language Models}, 
      author={Haoran Wei and Lingyu Kong and Jinyue Chen and Liang Zhao and Zheng Ge and Jinrong Yang and Jianjian Sun and Chunrui Han and Xiangyu Zhang},
      year={2023},
      eprint={2312.06109},
      archivePrefix={arXiv},
      primaryClass={cs.CV},
      url={https://arxiv.org/abs/2312.06109}, 
}

@misc{zhang2025biomedclipmultimodalbiomedicalfoundation,
      title={BiomedCLIP: a multimodal biomedical foundation model pretrained from fifteen million scientific image-text pairs}, 
      author={Sheng Zhang and Yanbo Xu and Naoto Usuyama and Hanwen Xu and Jaspreet Bagga and Robert Tinn and Sam Preston and Rajesh Rao and Mu Wei and Naveen Valluri and Cliff Wong and Andrea Tupini and Yu Wang and Matt Mazzola and Swadheen Shukla and Lars Liden and Jianfeng Gao and Angela Crabtree and Brian Piening and Carlo Bifulco and Matthew P. Lungren and Tristan Naumann and Sheng Wang and Hoifung Poon},
      year={2025},
      eprint={2303.00915},
      archivePrefix={arXiv},
      primaryClass={cs.CV},
      url={https://arxiv.org/abs/2303.00915}, 
}

@article{bubeck2023sparks,
  title={Sparks of artificial general intelligence: early experiments with GPT-4 (2023)},
  author={Bubeck, S{\'e}bastien and Chandrasekaran, Varun and Eldan, Ronen and Gehrke, Johannes and Horvitz, Eric and Kamar, Ece and Lee, Peter and Lee, Yin Tat and Li, Yuanzhi and Lundberg, Scott and others},
  journal={arXiv preprint arXiv:2303.12712},
  volume={1},
  year={2023}
}

@article{eloundou2023gpts,
  title={Gpts are gpts: An early look at the labor market impact potential of large language models},
  author={Eloundou, Tyna and Manning, Sam and Mishkin, Pamela and Rock, Daniel},
  journal={arXiv preprint arXiv:2303.10130},
  volume={10},
  year={2023}
}

@article{li2023ethics,
  title={Ethics of large language models in medicine and medical research},
  author={Li, Hanzhou and Moon, John T and Purkayastha, Saptarshi and Celi, Leo Anthony and Trivedi, Hari and Gichoya, Judy W},
  journal={The Lancet Digital Health},
  volume={5},
  number={6},
  pages={e333--e335},
  year={2023},
  publisher={Elsevier}
}

@book{bloom1956taxonomy,
  title={Taxonomy of educational objectives: The classification of educational goals. Handbook 1: Cognitive domain},
  author={Bloom, Benjamin S and others}
}

@inproceedings{carion2020end,
  title={End-to-end object detection with transformers},
  author={Carion, Nicolas and Massa, Francisco and Synnaeve, Gabriel and Usunier, Nicolas and Kirillov, Alexander and Zagoruyko, Sergey},
  booktitle={European conference on computer vision},
  pages={213--229},
  year={2020},
  organization={Springer}
}

@inproceedings{radford2021learning,
  title={Learning transferable visual models from natural language supervision},
  author={Radford, Alec and Kim, Jong Wook and Hallacy, Chris and Ramesh, Aditya and Goh, Gabriel and Agarwal, Sandhini and Sastry, Girish and Askell, Amanda and Mishkin, Pamela and Clark, Jack and others},
  booktitle={International conference on machine learning},
  pages={8748--8763},
  year={2021},
  organization={PmLR}
}

@misc{vicuna2023,
    title = {Vicuna: An Open-Source Chatbot Impressing GPT-4 with 90\%* ChatGPT Quality},
    url = {https://lmsys.org/blog/2023-03-30-vicuna/},
    author = {Chiang, Wei-Lin and Li, Zhuohan and Lin, Zi and Sheng, Ying and Wu, Zhanghao and Zhang, Hao and Zheng, Lianmin and Zhuang, Siyuan and Zhuang, Yonghao and Gonzalez, Joseph E. and Stoica, Ion and Xing, Eric P.},
    month = {March},
    year = {2023}
}

@article{zhou2025perception,
  title={From perception to cognition: A survey of vision-language interactive reasoning in multimodal large language models},
  author={Zhou, Chenyue and Wang, Mingxuan and Ma, Yanbiao and Wu, Chenxu and Chen, Wanyi and Qian, Zhe and Liu, Xinyu and Zhang, Yiwei and Wang, Junhao and Xu, Hengbo and others},
  journal={arXiv preprint arXiv:2509.25373},
  year={2025}
}

@article{li2025perception,
  title={Perception, reason, think, and plan: A survey on large multimodal reasoning models},
  author={Li, Yunxin and Liu, Zhenyu and Li, Zitao and Zhang, Xuanyu and Xu, Zhenran and Chen, Xinyu and Shi, Haoyuan and Jiang, Shenyuan and Wang, Xintong and Wang, Jifang and others},
  journal={arXiv preprint arXiv:2505.04921},
  year={2025}
}

@inproceedings{liu2024mmbench,
  title={Mmbench: Is your multi-modal model an all-around player?},
  author={Liu, Yuan and Duan, Haodong and Zhang, Yuanhan and Li, Bo and Zhang, Songyang and Zhao, Wangbo and Yuan, Yike and Wang, Jiaqi and He, Conghui and Liu, Ziwei and others},
  booktitle={European conference on computer vision},
  pages={216--233},
  year={2024},
  organization={Springer}
}

@inproceedings{singh2019towards,
  title={Towards vqa models that can read},
  author={Singh, Amanpreet and Natarajan, Vivek and Shah, Meet and Jiang, Yu and Chen, Xinlei and Batra, Dhruv and Parikh, Devi and Rohrbach, Marcus},
  booktitle={Proceedings of the IEEE/CVF conference on computer vision and pattern recognition},
  pages={8317--8326},
  year={2019}
}

@inproceedings{li2023evaluating,
  title={Evaluating object hallucination in large vision-language models},
  author={Li, Yifan and Du, Yifan and Zhou, Kun and Wang, Jinpeng and Zhao, Wayne Xin and Wen, Ji-Rong},
  booktitle={Proceedings of the 2023 conference on empirical methods in natural language processing},
  pages={292--305},
  year={2023}
}

@article{kargupta2025cognitive,
  title={Cognitive foundations for reasoning and their manifestation in llms},
  author={Kargupta, Priyanka and Li, Shuyue Stella and Wang, Haocheng and Lee, Jinu and Chen, Shan and Ahia, Orevaoghene and Light, Dean and Griffiths, Thomas L and Kleiman-Weiner, Max and Han, Jiawei and others},
  journal={arXiv preprint arXiv:2511.16660},
  year={2025}
}

@article{lu2024ai,
  title={The ai scientist: Towards fully automated open-ended scientific discovery},
  author={Lu, Chris and Lu, Cong and Lange, Robert Tjarko and Foerster, Jakob and Clune, Jeff and Ha, David},
  journal={arXiv preprint arXiv:2408.06292},
  year={2024}
}

@article{bai2023qwen,
  title={Qwen technical report},
  author={Bai, Jinze and Bai, Shuai and Chu, Yunfei and Cui, Zeyu and Dang, Kai and Deng, Xiaodong and Fan, Yang and Ge, Wenbin and Han, Yu and Huang, Fei and others},
  journal={arXiv preprint arXiv:2309.16609},
  year={2023}
}

@article{dai2023instructblip,
  title={Instructblip: Towards general-purpose vision-language models with instruction tuning},
  author={Dai, Wenliang and Li, Junnan and Li, Dongxu and Tiong, Anthony and Zhao, Junqi and Wang, Weisheng and Li, Boyang and Fung, Pascale N and Hoi, Steven},
  journal={Advances in neural information processing systems},
  volume={36},
  pages={49250--49267},
  year={2023}
}

@article{wijaya2024multimodal,
  title={Multimodal preference data synthetic alignment with reward model},
  author={Wijaya, Robert and Nguyen, Ngoc-Bao and Cheung, Ngai-Man},
  journal={arXiv e-prints},
  pages={arXiv--2412},
  year={2024}
}

@article{yu2023mm,
  title={Mm-vet: Evaluating large multimodal models for integrated capabilities},
  author={Yu, Weihao and Yang, Zhengyuan and Li, Linjie and Wang, Jianfeng and Lin, Kevin and Liu, Zicheng and Wang, Xinchao and Wang, Lijuan},
  journal={arXiv preprint arXiv:2308.02490},
  year={2023}
}

@article{li2023seed,
  title={Seed-bench: Benchmarking multimodal llms with generative comprehension},
  author={Li, Bohao and Wang, Rui and Wang, Guangzhi and Ge, Yuying and Ge, Yixiao and Shan, Ying},
  journal={arXiv preprint arXiv:2307.16125},
  year={2023}
}

@inproceedings{liu2024improved,
  title={Improved baselines with visual instruction tuning},
  author={Liu, Haotian and Li, Chunyuan and Li, Yuheng and Lee, Yong Jae},
  booktitle={Proceedings of the IEEE/CVF conference on computer vision and pattern recognition},
  pages={26296--26306},
  year={2024}
}

@misc{li2024llavanextinterleavetacklingmultiimagevideo,
      title={LLaVA-NeXT-Interleave: Tackling Multi-image, Video, and 3D in Large Multimodal Models}, 
      author={Feng Li and Renrui Zhang and Hao Zhang and Yuanhan Zhang and Bo Li and Wei Li and Zejun Ma and Chunyuan Li},
      year={2024},
      eprint={2407.07895},
      archivePrefix={arXiv},
      primaryClass={cs.CV},
      url={https://arxiv.org/abs/2407.07895}, 
}
\clearpage
\appendix
\section{Appendix / Supplementary Materials}
\subsection{Text Evidence}
The resulting evidence annotations also allow us to examine text patterns within each dataset. Table~\ref{tab:textvqa_text_evidence_token_buckets} shows the distribution of text-token counts in TextVQA, while Table~\ref{tab:mme_text_evidence_token_buckets} presents the corresponding distribution for MME. Additional summary statistics of text-token counts for both TextVQA and MME are provided in Table~\ref{tab:text-atomicity-compare}.

\begin{table}[H]
\centering

\begin{subtable}[t]{0.48\textwidth}
\centering
\caption{Distribution of tokens on TextVQA.}
\label{tab:textvqa_text_evidence_token_buckets}
\begin{tabular}{lrr}
\toprule
Tokens in \texttt{TEXT} & Count & Percentage \\
\midrule
1      & 323  & 6.6\% \\
2--5   & 1426 & 29.2\% \\
6--10  & 1094 & 22.4\% \\
11--20 & 1112 & 22.8\% \\
21--32 & 468  & 9.6\% \\
33--50 & 258  & 5.3\% \\
$>50$  & 204  & 4.2\% \\
\bottomrule
\end{tabular}
\end{subtable}
\hfill
\begin{subtable}[t]{0.48\textwidth}
\centering
\caption{Distribution of tokens on MME.}
\label{tab:mme_text_evidence_token_buckets}
\begin{tabular}{lrr}
\toprule
Tokens in \texttt{TEXT} & Count & Percentage \\
\midrule
1      & 22  & 1.5\% \\
2--5   & 122 & 8.1\% \\
6--10  & 170 & 11.3\% \\
11--20 & 224 & 14.9\% \\
21--32 & 240 & 15.9\% \\
33--50 & 237 & 15.7\% \\
$>50$  & 492 & 32.6\% \\
\bottomrule
\end{tabular}
\end{subtable}

\caption{Distribution of the number of tokens in \texttt{TEXT} evidence.}
\label{tab:text_evidence_token_buckets_compare}
\end{table}
\vspace{-3em}
\begin{table}[H]
\centering
\small
\begin{tabular}{lcc}
\toprule
\textbf{TEXT evidence atomicity statistic} & \textbf{TextVQA} & \textbf{MME} \\
\midrule
Number of questions with \texttt{TEXT} evidence ($N$) & 4885 & 1507 \\
Mean quoted OCR tokens per \texttt{TEXT} item & 13.5 & 47.3 \\
Median quoted OCR tokens per \texttt{TEXT} item & 8 & 31 \\
90th percentile (p90) quoted OCR tokens per \texttt{TEXT} item & 32 & 103 \\
Maximum quoted OCR tokens per \texttt{TEXT} item & 95 & 522 \\
$\Pr(\text{token count} > 1)$ & 93.4\% & 98.5\% \\
$\Pr(\text{token count} > 10)$ & 41.8\% & 79.2\% \\
\bottomrule
\end{tabular}
\caption{Atomicity of \texttt{TEXT} evidence on TextVQA and MME.}
\label{tab:text-atomicity-compare}
\end{table}
\vspace{-4em}
\subsection{Bloom Verbalization Examples}
\begin{figure}[tb]
  \centering
  \includegraphics[width=\linewidth, height=13cm, keepaspectratio]{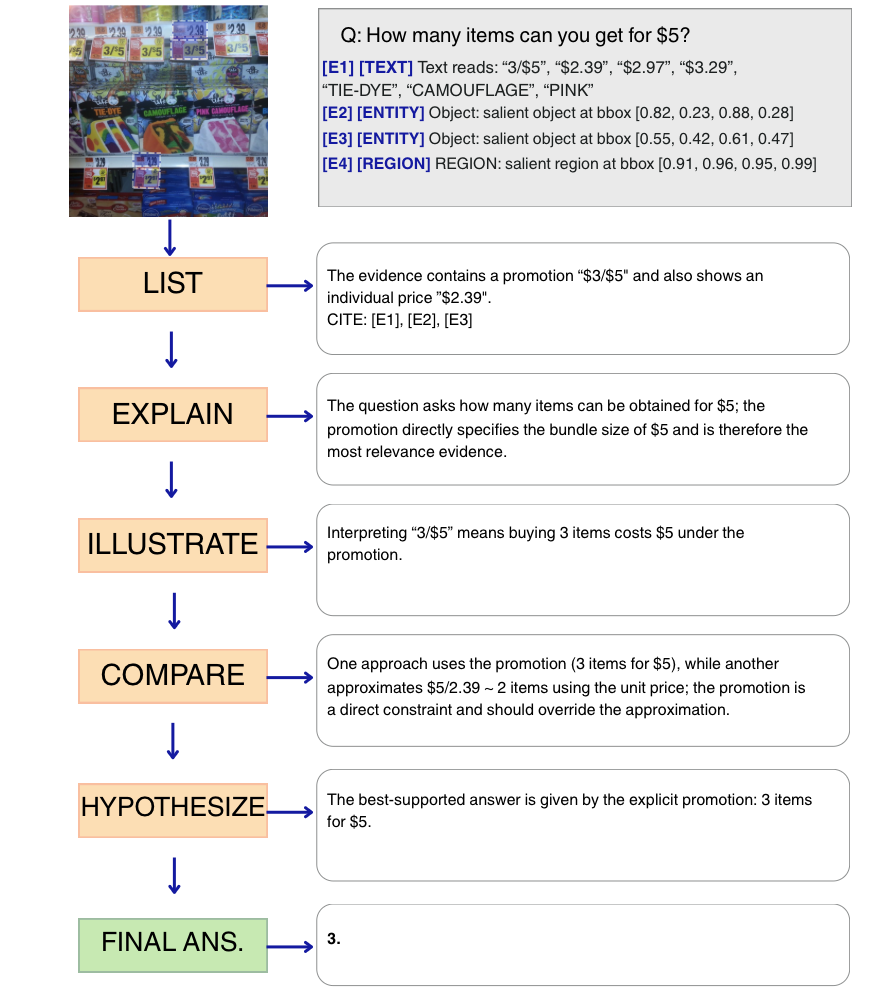}
  \caption{An example to illustrate the Bloom verbalization (LIST to Final Answer) reasoning progression for determining number of items can buy with certain amount.}
  \label{fig:appendix5}
\end{figure}

\begin{figure}[tb]
  \centering
  \includegraphics[width=\linewidth, height=13cm, keepaspectratio]{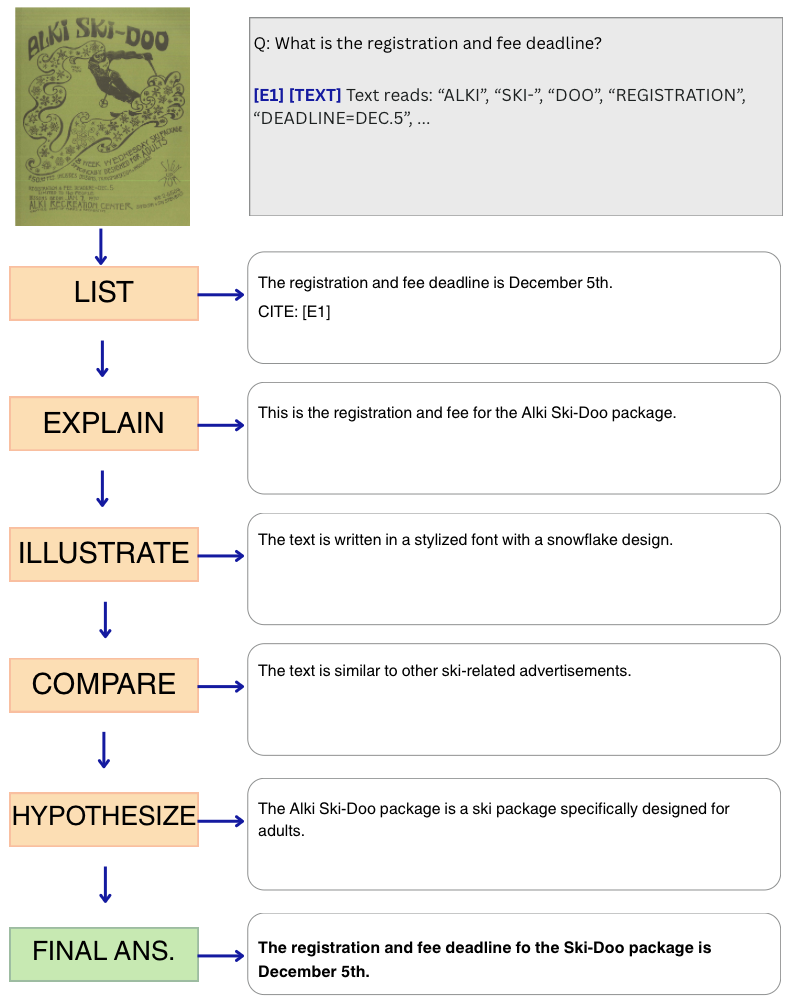}
  \caption{An example to illustrate the Bloom verbalization (LIST to Final Answer) reasoning progression for recognition \& OCR.}
  \label{fig:appendix6}
\end{figure}

\begin{figure}[tb]
  \centering
  \includegraphics[width=\linewidth, height=13cm, keepaspectratio]{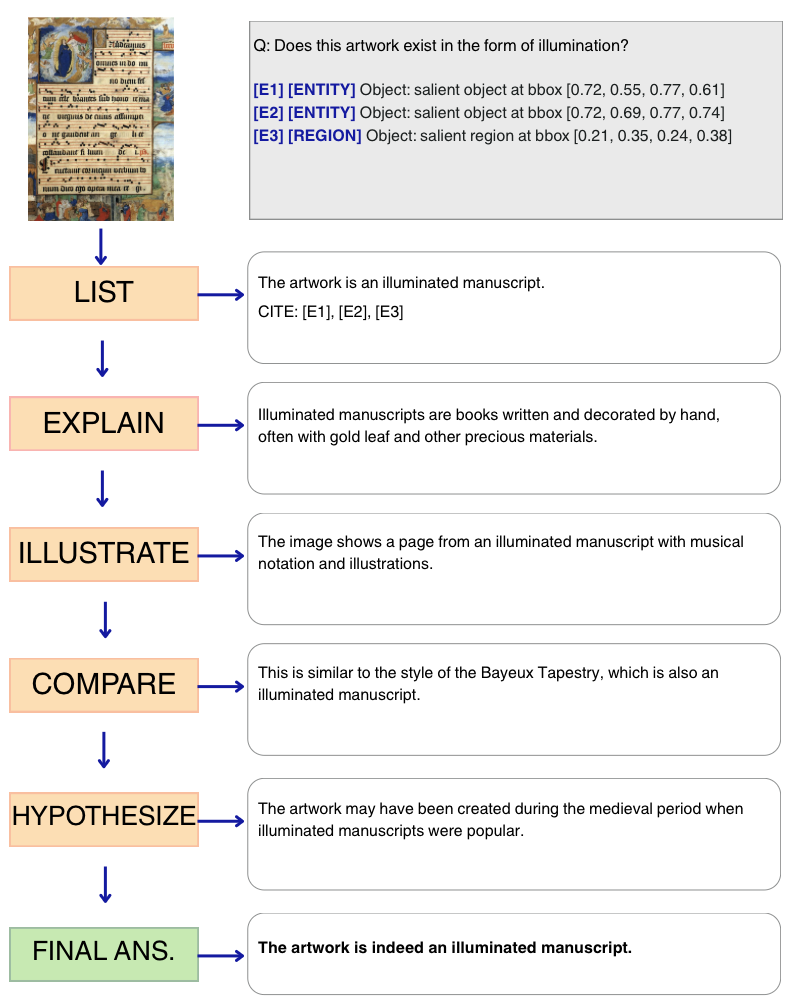}
  \caption{Another example to illustrate the Bloom verbalization (LIST to Final Answer) reasoning progression in the topic of artwork.}
  \label{fig:appendix7}
\end{figure}

We provide additional examples of Bloom verbalization steps in this appendix. Figure~\ref{fig:appendix5} presents an example involving numerical calculation, Figure~\ref{fig:appendix6} shows an OCR-based task, and Figure~\ref{fig:appendix7} illustrates a perception task on artwork.

\newcolumntype{L}[1]{>{\raggedright\arraybackslash}p{#1}}
\newcolumntype{Y}{>{\raggedright\arraybackslash}X}

\renewcommand{\arraystretch}{1.35}

\begin{table}[t]
\centering
\setlength{\tabcolsep}{10pt}

\begin{tabularx}{0.86\linewidth}{@{}L{0.20\linewidth}Y@{}}
\Xhline{1.2pt}
Input Image: &
\begin{minipage}[t]{\linewidth}
\vspace{0.2em}
\includegraphics[width=0.50\linewidth]{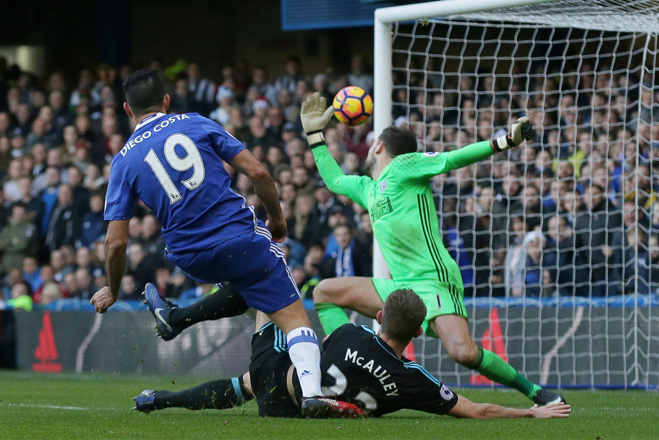}
\vspace{0.1em}
\end{minipage}
\\
\hline
User &
\makecell[l]{How many soccer players are present in the image?\\
A. 3 \quad B. 2 \quad C. 1 \quad D. 4}
\\
\hline
InstructBLIP & D. 4 \\
\hline
Qwen-VL & D. 4 \\
\hline
LLaVA1.5 7B & D. 4 \\
\hline
LLaVA1.5 13B & A. 3 \\
\hline
MoCE & A. 3 \\
\Xhline{1.2pt}
\end{tabularx}

\vspace{0.6em}

\begin{tabularx}{0.86\linewidth}{@{}L{0.20\linewidth}Y@{}}
\Xhline{1.2pt}
Input Image: &
\begin{minipage}[t]{\linewidth}
\par
\vspace{0.2em}
\includegraphics[width=0.50\linewidth]{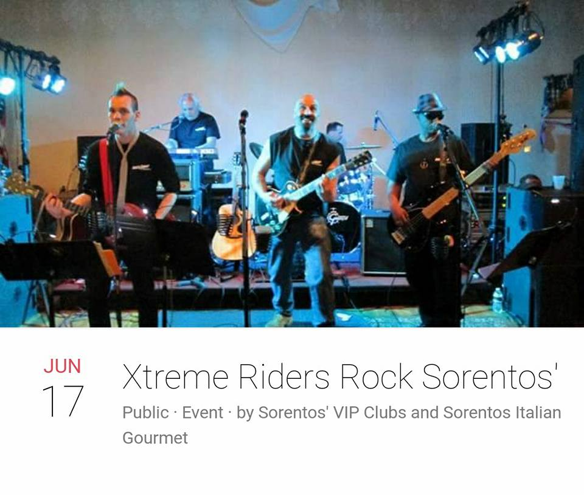}
\vspace{0.1em}
\end{minipage}
\\
\hline
User &
\makecell[l]{What is the action of the person wearing the hat \\on stage?\\
A. Playing the guitar \quad B. Singing \quad C. Dancing \quad\\ D. Playing the drums}
\\
\hline
InstructBLIP & A. Playing the guitar \\
\hline
Qwen-VL & D. Playing the drums \\
\hline
LLaVA1.5 7B & D. Playing the drums \\
\hline
LLaVA1.5 13B & A. Playing the guitar \\
\hline
MoCE & A. Playing the guitar \\
\Xhline{1.2pt}
\end{tabularx}

\caption{Qualitative examples on SEED-IMG.}
\label{tab:seed_grouped_examples}
\end{table}

\subsection{JSON File Format}
\begin{figure}[tb]
  \centering
  \includegraphics[width=\linewidth, height=14cm, keepaspectratio]{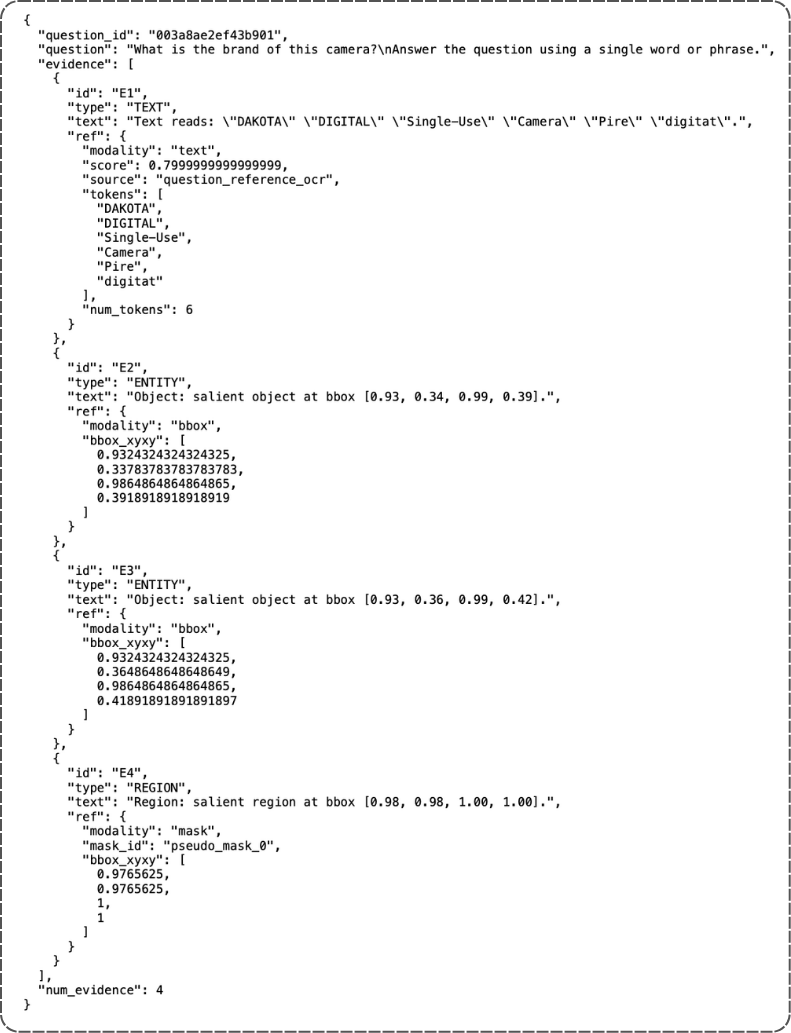}
  \caption{JSON schema of the evidences.}
  \label{fig:appendix8}
\end{figure}

The generated evidence is stored in JSON format. As illustrated in Fig.~\ref{fig:appendix8}, the JSON schema first presents the question, followed by the detailed evidence entries.

\subsection{Qualitative Assessment}
\begin{table}[H]
\centering
\setlength{\tabcolsep}{10pt}

\begin{tabularx}{0.86\linewidth}{@{}L{0.20\linewidth}Y@{}}
\Xhline{1.2pt}
Input Image: &
\begin{minipage}[t]{\linewidth}
\vspace{0.2em}
\includegraphics[width=0.50\linewidth]{figures/appendix3.jpg}
\vspace{0.1em}
\end{minipage}
\\
\hline
User &
\makecell[l]{On the right desk, what is to the left of the laptop?}
\\
\hline
InstructBLIP & Printer \\
\hline
Qwen-VL & Printer \\
\hline
LLaVA1.5 7B & Books \\
\hline
LLaVA1.5 13B & A lamp \\
\hline
MoCE & A lamp \\
\Xhline{1.2pt}
\end{tabularx}

\vspace{0.6em}

\begin{tabularx}{0.86\linewidth}{@{}L{0.20\linewidth}Y@{}}
\Xhline{1.2pt}
Input Image: &
\begin{minipage}[t]{\linewidth}
\par
\vspace{0.2em}
\includegraphics[width=0.50\linewidth]{figures/appendix4.png}
\vspace{0.1em}
\end{minipage}
\\
\hline
User &
\makecell[l]{What is the price difference between diesel and \\regular gasoline?\\}
\\
\hline
InstructBLIP & 0 \\
\hline
Qwen-VL & 0.90 \\
\hline
LLaVA1.5 7B & 0.36 \\
\hline
LLaVA1.5 13B & 0.30 \\
\hline
MoCE & 0.11 \\
\Xhline{1.2pt}
\end{tabularx}

\caption{Qualitative examples on MM-Vet.}
\label{tab:mmvet_grouped_examples}
\end{table}

\end{document}